\documentclass[letterpaper, 10 pt, conference]{ieeeconf}  % Comment this line out if you need a4paper

\IEEEoverridecommandlockouts                              % This command is only needed if 
\usepackage[linesnumbered,ruled,vlined]{algorithm2e}
\usepackage{algorithmic}
\usepackage{graphics} % for pdf, bitmapped graphics files
\graphicspath{{figures/}}
\usepackage{epsfig} % for postscript graphics files
\usepackage{times} % assumes new font selection scheme installed
\usepackage{siunitx}
\usepackage{amsmath} % assumes amsmath package installed
\usepackage{amssymb}  % assumes amsmath package installed
\usepackage{hyperref}
\usepackage{multirow}
\usepackage[capitalise]{cleveref}
\Crefname{algorithm}{Alg.}{Algs.}
\Crefname{section}{Sec.}{Secs.}
\Crefname{equation}{Eq.}{Eqs.}
\usepackage[colorinlistoftodos]{todonotes}
\usepackage{booktabs}
\usepackage[
    sorting=none,
    giveninits=true,
    maxbibnames=9,
    maxcitenames=2,backend=biber
    ]{biblatex}
\title{\LARGE \bf
A Data-Efficient Approach for Long-Term Human Motion Prediction Using Maps of Dynamics
}

\author{
Yufei Zhu$^{1}$, Andrey Rudenko$^{2}$, Tomasz Piotr Kucner$^{3}$, Achim J. Lilienthal$^{1,4}$, Martin Magnusson$^{1}$%
\thanks{$^{1}$AASS MRO lab, Örebro University, Sweden {\tt\small yufei.zhu@oru.se}}%
\thanks{$^{2}$Bosch Corporate Research, Robert Bosch GmbH, Stuttgart, Germany {\tt\small andrey.rudenko@bosch.com}}%
\thanks{$^{3}$Aalto University, Finland}%
\thanks{$^{4}$Chair ``Perception for Intelligent Systems'', TU Munich, Germany}%
\thanks{This work has received funding from the European Union’s Horizon 2020 research and innovation programme under grant agreement No 101017274 (DARKO).}
}

\begin{document}

\maketitle
\thispagestyle{empty}
\pagestyle{empty}

%%%%%%%%%%%%%%%%%%%%%%%%%%%%%%%%%%%%%%%%%%%%%%%%%%%%%%%%%%%%%%%%%%%%%%%%%%%%%%%%
\begin{abstract}

% workshop version
% Focus on data efficient
Human motion prediction is essential for the safe and smooth operation of mobile service robots and intelligent vehicles around people. Commonly used neural network-based approaches often require large amounts of complete trajectories to represent motion dynamics in complex semantically-rich spaces. This requirement may complicate deployment of physical systems in new environments, especially when the data is being collected online from onboard sensors. In this paper we explore a data-efficient alternative using \emph{maps of dynamics} (MoD) to represent place-dependent multi-modal spatial motion patterns, learned from prior observations. Our approach can perform efficient human motion prediction in the long-term perspective of up to 60 seconds. We quantitatively evaluate its accuracy with limited amount of training data in comparison to an LSTM-based baseline, and qualitatively show that the predicted trajectories reflect the natural semantic properties of the environment, e.g. the locations of short- and long-term goals, navigation in narrow passages, around obstacles, etc.

\end{abstract}

%%%%%%%%%%%%%%%%%%%%%%%%%%%%%%%%%%%%%%%%%%%%%%%%%%%%%%%%%%%%%%%%%%%%%%%%%%%%%%%%

\section{INTRODUCTION}

Long-term human motion prediction (LHMP) is important for autonomous robots and vehicles to operate safely in populated environments \cite{rudenko2020human}. Accurately predicting the future trajectories of people in their surroundings over extended time periods is essential for enhancing motion planning, tracking, automated driving, human-robot interaction, intelligent safety monitoring and surveillance.

Human motion is complex and may be influenced by several hard-to-model factors, including social rules and norms, personal preferences, and subtle cues in the environment that are not represented in geometric maps. To address these challenges, popular neural network approaches learn motion dynamics directly from data, with many recent studies developing models based on LSTMs \cite{alahi2016social}, GANs \cite{sadeghian2018sophie}, CNNs \cite{mohamed2020social}, CVAEs \cite{salzmann2020trajectronpp} and transformers \cite{giuliari2021transformer}. Most of these approaches focus on learning to predict stochastic interactions between diverse moving agents in the short-term perspective in scenarios where the effect of the environment topology and semantics is minimal.

When predicting long-term human motion in complex, large-scale environments, the influence of the surrounding space (e.g. passages, stairs, entrances,  various objects and semantically-meaningful areas) on human motion goes beyond what is contained in the current state of the moving person or the observed interactions. This impact of the environment has to be modelled explicitly, for instance by informing the prediction method with a semantic map \cite{wuIV18,zhao2019multi,rudenko2020semapp}. Another effective approach to address this challenge is to use \emph{maps of dynamics} (MoDs). MoDs \cite{kucner2020probabilistic} are maps that encode spatial or spatio-temporal motion patterns as a feature of the environment. 
%By building spatial and spatio-temporal models, MoDs capture the patterns followed by dynamic objects in the environment. 
MoD-informed long-term human motion prediction (MoD-LHMP) approaches are particularly suited to predict motion in the long-term perspective, where the environment effects become critical for making accurate predictions. MoDs efficiently encode the stochastic local motion patterns over the entire map, informing the predictor in areas which may have no influence on the immediate decisions of the walking people, but become critical in the long-term perspective. 

As a proof of concept for MoD-LHMP, we propose to build CLiFF MoDs \cite{kucner2017enabling} from training data and use them to bias a constant velocity motion prediction method, generating stochastic trajectory predictions for up to \SI{60}{\second} into the future. % Exploiting human motion patterns captured by MoDs, prior work proposed an approach named CLiFF-LHMP, which uses one type of MoDs, CLiFF-map to predict long-term human motion up to \SI{50}{\second}. CLiFF-LHMP biases a constant velocity prediction with samples from CLiFF-map to generate multi-modal trajectory predictions. 

%With exploiting other types of MoDs, CLiFF-LHMP can be extended to MoD-LHMP, a class of MoD-informed long-term human motion prediction approaches. For MoDs used for LHMP, besides CLiFF-map, other MoDs based on velocity observations can be applied. One option is STeF-map, which incorporated temporal information when building flow models. In STeF-map, \textcite{molina2018modelling} apply the Frequency Map Enhancement (FreMEn \cite{krajnik2017fremen}), which is a model describing spatiotemporal dynamics in the frequency domain, to build a time-dependent probabilistic map to model periodic changes in people flow. Unlike CliFF-map, where speed and orientation are represented as a continuous joint distribution, the orientation is discretized in STeF-map.

\begin{figure}[t]
\centering
\includegraphics[width=.48\linewidth]{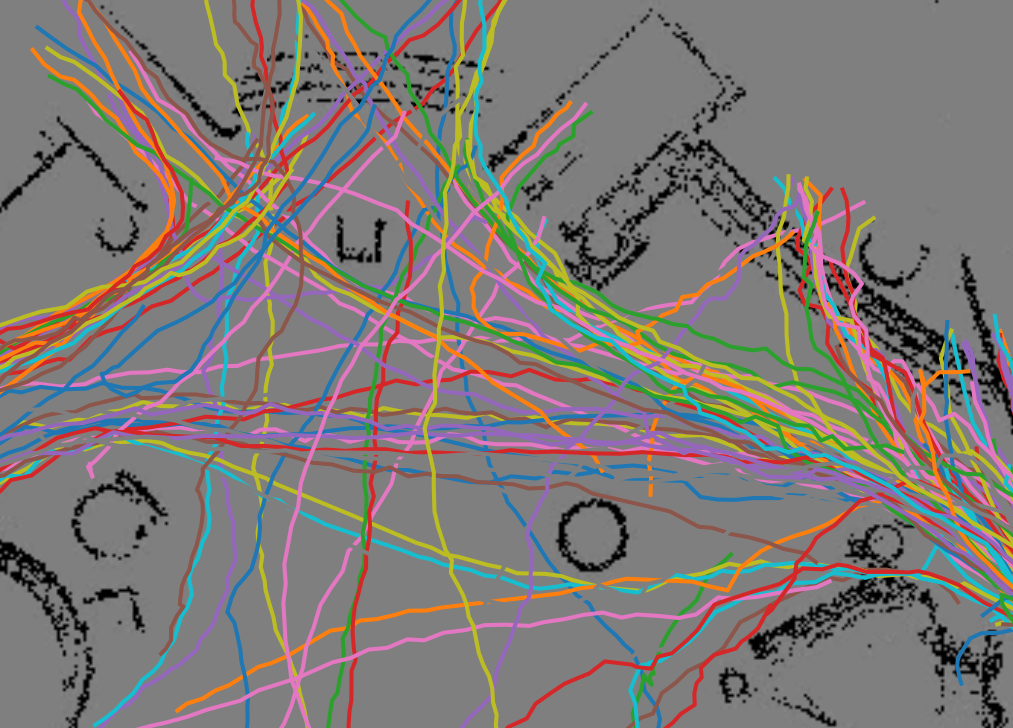}
\includegraphics[width=.48\linewidth]{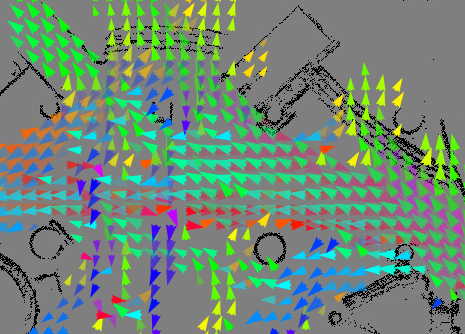}
\caption{Maps of dynamics provide an efficient and lightweight encoding of sparse and incomplete velocity data to characterize the motion flows in the environment. We propose a method to predict long-term multi-modal human motion using data-efficient CLiFF maps \cite{kucner2017enabling}. {\bf Left:} trajectories from the ATC dataset used for training. {\bf Right:} CLiFF map.}
\label{fig:cover}
\vspace*{-3mm}
\end{figure}

One crucial advantage of the MoD-LHMP approach is its data efficiency. Prior art neural network-based approaches often require large amounts of data for training, and their performance can significantly degrade in absence thereof. Typically, these approaches also need complete sequences of tracked positions for training. The proposed MoD-LHMP approach, on the other hand, allows encoding human motion from sparse and incomplete data, requiring only observed velocities in discrete locations and interpolating the missing motion in between. This property is relevant, for instance, when the deployed robot collects the data in an online fashion from on-board sensors and with a limited field of view.

%approach offers the advantage of data efficiency, requiring minimal amounts of data to achieve accurate predictions. In this work, we evaluate the data-efficient characteristic of MoD-based approaches.

In this work, we evaluate the efficiency of MoD-based motion encoding for making accurate long-term predictions. In our experiments we sample few trajectories from the ATC dataset and use them to build a CLiFF map and train LSTM-based baselines. We then compare these methods using the ADE/FDE prediction accuracy metrics. Furthermore, we qualitatively demonstrate that the CLiFF-LHMP approach has the ability to predict human motion in complex environments over very long time horizons, implicitly inferring common goal points and correctly predicting trajectories that follow the complex topology of the environment, e.g. navigating around corners or obstacles or passing through narrow passages such as doors.

\section{METHOD} \label{section-method}

\subsection{Maps of Dynamics}

In the proposed approach for human motion prediction, we exploit Maps of Dynamics (MoD) which encode human dynamics as a feature of the environment. By using velocity observations, human dynamics can be represented through flow models. In this work, we employ Circular-Linear Flow Field map (CLiFF-map) \cite{kucner2017enabling} to represent the flow of human motion. CLiFF-map represents local flow patterns as a multi-modal, continuous joint distribution of speed and orientation. As the orientation of velocity is a circular variable, and magnitude of velocity is a linear variable, CLiFF-map associates a semi-wrapped Gaussian mixture model (SWGMM) with each location, describing flow patterns around the given location, see Fig.~\ref{fig:cover}. By using SWGMM, CLiFF-map is able to properly address multimodality in the data, thereby enhancing its capability to predict uncertain long-term human motion. A CLiFF-map represents motion patterns based on local observations and estimates the likelihood of motion at a given query location. As it can be built from incomplete or spatially sparse data, CLiFF-map efficiently captures human motion patterns without requiring large amounts of data or complete trajectories. This characteristic makes CLiFF-LHMP a data-efficient approach for predicting human motion.

\subsection{Motion Prediction}

We frame the task of predicting a person's future trajectory as using a short observed trajectory to infer a sequence of future states. The length of the observation history is $O_s \in \mathbb{R}^+$ \SI{}{\second}, equivalent to an integer $O_p > 0$ observation time steps. With the current time-step denoted as the integer $t_0 \geq 0$, %$t_0 \in \mathbb{N}$, 
the sequence of observed states is $\mathcal{H} = \langle s_{t_{0} - 1},..., s_{t_{0} - O_p} \rangle$, where $s_t$ is the state of a person at time-step $t$. 
%Referring to a world frame, 
A state is represented by 2D Cartesian coordinates $(x,y)$, speed $\rho$ and orientation $\theta$:
%\begin{equation}
$s = (x,y,\rho, \theta)$.
%\end{equation}

From the observed sequence $\mathcal{H}$, we derive the observed speed $\rho_{\mathrm{obs}}$ and orientation $\theta_{\mathrm{obs}}$ at time-step $t_0$. Then the current state becomes $s_{t_0} = (x_{t_0},y_{t_0},\rho_{\mathrm{obs}},\theta_{\mathrm{obs}})$. The values of $\rho_{\mathrm{obs}}$ and  $\theta_{\mathrm{obs}}$ are calculated as a weighted sum of the finite differences in the observed states, as in the recent ATLAS benchmark \cite{rudenko2022atlas}, such that $\rho_{\mathrm{obs}} = \sum_{t=1}^{O_p}v_{t_0 - t}g(t)$ and $\theta_{\mathrm{obs}} = \sum_{t=1}^{O_p}\theta_{t_0 - t}g(t)$, where $g(t) = %\frac{1}{\sigma\sqrt{2\pi}e^{-\frac{t^2}{\sigma}}}$.
%(\sigma\sqrt{2\pi}e^{-\frac{t^2}{\sigma}})^{-1}$.
(\sigma\sqrt{2\pi}e^{\frac{1}{2}(\frac{t}{\sigma})^2})^{-1}$.

Given the current state $s_{t_0}$, the goal is to estimate a sequence of future states.
%Similar to past states, future
Future
states are predicted for a given horizon $T_s \in \mathbb{R}^+$ \SI{}{\second}. $T_s$ is equivalent to 
$T_p > 0$ %$T_p \in \mathbb{Z}^+$ 
prediction time steps assuming the constant time interval $\Delta t$ between two predictions. Thus, the prediction horizon is $T_s = T_p \Delta t$. The future sequence is then denoted as $\mathcal{T} = \langle s_{t_0+1}, s_{t_0+2},...,s_{t_0+T_p} \rangle$.

% To predict human trajectories we exploit the information about local flow patterns represented in a CLiFF-map as a multimodal, continuous distribution over oriented velocities. CLiFF-map~\cite{kucner2017enabling} is a probabilistic framework for mapping velocity observations (independently of their underlying physical processes), i.e., essentially a generalization of a vector field into a Gaussian mixture field. Each location in the map is associated with a Gaussian mixture model (GMM) and variables describing the confidence in each model. A CLiFF-map represents motion patterns based on local observations and estimates the likelihood of motion at a given query location. 

\begin{algorithm}[t]
\small
    \KwIn{$\mathcal{H}$, $x_{t_0}$, $y_{t_0}$}
    \KwOut{$\mathcal{T}$}
	%\SetAlgoLined
	$\mathcal{T} = \{\}$ \
	
	$\rho_\mathrm{obs}, \theta_\mathrm{obs} \leftarrow $  getObservedVelocity($\mathcal{H}$) \
	
	$s_{t_0} = (x_{t_0},y_{t_0},\rho_{\mathrm{obs}},\theta_{\mathrm{obs}})$ \
	
	\For { $t= t_{0}+1$, ..., $t_{0}+T_p$}{
	
	    $x_t, y_t \leftarrow $ getNewPosition($s_{t\textendash1}$) \
	    
		% $\xi \leftarrow $ selectSWGMM($x_t, y_t$)\
	    
		$\theta_s$ $ \leftarrow $ sampleVelocityFromCLiFFmap($x_t, y_t$)\
	    
		($\rho_t$, $\theta_t$) $ \leftarrow $ predictVelocity($\theta_{s}$, $\rho_{t\textendash1}$, $\theta_{t\textendash1}$)\
		
		$s_t \leftarrow (x_t, y_t, \rho_{t}, \theta_t)$
		
		$\mathcal{T} \leftarrow \mathcal{T} \cup s_t$ \
    }
    \Return $\mathcal{T}$ \

 %\caption{Motion prediction using CVM and MoD}
 \caption{CLiFF-LHMP}
\label{alg:LHMPAlgo}
\end{algorithm}

The CLiFF-LHMP algorithm is presented in \cref{alg:LHMPAlgo}. With the input of a CLiFF-map and past states of a person, the algorithm predicts a sequence of future states. To estimate $\mathcal{T}$, for each prediction time step, we sample a velocity from the CLiFF-map at the current position ($x_t$, $y_t$) to bias the prediction with the learned motion patterns represented by the CLiFF-map. %One example of the process of sampling a predicted motion state from a CLiFF-map is illustrated in \cref{fig:sample}. 
To sample a velocity at a given location $(x,y)$, we first get the SWGMMs $\Xi_{\mathrm{near}}$ whose distances to $(x,y)$ are less than $r_s$, where $r_s$ is the sampling radius. After getting the sampled velocity, the velocity ($\rho_t$, $\theta_t$) is predicted by assuming that a person will continue walking with the same speed as in the last time step, $\rho_t = \rho_{t-1}$, and biasing the direction of motion with the sampled orientation $\theta_s$ as:

\begin{equation}
\begin{gathered}
\theta_t = \theta_{t-1} + (\theta_s - \theta_{t-1}) \cdot K(\theta_{s} - \theta_{t-1}), \\
\end{gathered}
\end{equation}

\noindent where $K(\cdot)$ is a kernel function that defines the degree of impact of the CLiFF-map. We use a Gaussian kernel with a parameter $\beta$ that represents the kernel width:
\begin{equation} \label{eq-kernel}
    %K(\theta_s - \theta_{t-1}) =  e ^ {-\beta \left\Vert \theta_s - \theta_{t-1} \right\Vert ^ 2}. \\
    K(x) =  e ^ {-\beta \left\Vert x \right\Vert ^ 2}. \\
\end{equation}

With kernel $K$, we scale the CLiFF-map term by the difference between the velocity sampled from the CLiFF-map and the current velocity according to a constant velocity model (CVM). The sampled velocity is trusted less if it deviates more from the current velocity.
A larger $\beta$ value makes the method behave more like a CVM, and a smaller $\beta$ makes it more closely follow the CLiFF-map.

%\begin{figure*}[t]
%\centering
%\includegraphics[width=.22\linewidth]{figures/sample1.png}
%\includegraphics[width=.22\linewidth]{figures/sample2.png}
%\includegraphics[width=.25\linewidth]{figures/cylinder1.pdf}%
%\includegraphics[width=.28\linewidth]{figures/sample4.png} \\
%\includegraphics[width=.80\linewidth]{figures/sub-figure-captions.pdf}

%\caption{Steps of sampling an orientation $\theta_s$ from the CLiFF-map. {\bf (a)} CLiFF-map built from the ATC data. The location to sample from is marked with an orange arrow. {\bf (b)} Selection of SWGMM in the CLiFF-map. The red circle contains all SWGMMs closer to the sample location than $r_s$. From these SWGMMs, the SWGMM with the highest motion ratio is selected (marked with a blue circle). {\bf (c)} The SWGMM distribution at the selected location. The distribution is wrapped on a unit cylinder. The speed is represented by the position along the $\rho$ axis and the orientation is denoted by $\theta$. The probability is represented by the distance of the surface to the axis of the cylinder. A velocity vector (marked with a red arrow) is sampled from this SWGMM. {\bf (d)} The orientation $\theta_s$ of this sampled velocity is shown in an orange circle.} \label{fig:sample}
%\vspace*{-6mm}
%\end{figure*}

\section{Evaluation} \label{section-experiments}
In this section, we evaluate the data efficiency and accuracy of the proposed CLiFF-LHMP approach and compare it to the LSTM-based human motion prediction methods. Vanilla LSTM \cite{hochreiter1997lstm} is used as the baseline representative of LSTM-based methods.

\subsection{Implementation Details}
We evaluate the prediction performance using the ATC dataset \cite{brscic2013person}, which contains trajectories recorded in a shopping mall in Japan. The dataset covers a large indoor environment with a total area of around \SI{900}{\metre\squared}. The ATC dataset consists of 92 days in total. Given the immense length of the ATC dataset for each recording day, a subset covering the first four days can be considered representative. We use the subset in the experiments, with the first day (Oct.24) for training, and the remaining 3 days for testing. Both the LSTM and CLiFF-LHMP approaches are trained with same data and evaluated with same data to ensure a fair comparison.

In ATC dataset, the original detection rate is \SI{30}{\Hz}. We downsample the data to \SI{2.5}{\Hz} to align with \SI{0.4}{\second} observation time interval, as commonly used in human motion prediction. For each trajectory, we take \SI{3.2}{\second} (the first 8 positions) as the observation history and use the remaining trajectory (up to the maximum prediction horizon) as the prediction ground truth. Instead of using a fixed prediction horizon, we explore a wider range of values $T_s$ up to a maximum value in our evaluation. The maximum prediction horizon is determined based on the tracking duration distribution of the dataset. We use the 90th percentile value, which is \SI{60}{\second}, as maximum prediction horizon for experiments of ATC dataset. As LSTM-based approaches require complete trajectories for training, we use for all compared approaches trajectories equal or longer than \SI{60}{\second} for both training and testing.

Given the area %($\sim$\SI{900}{\metre\squared})
and tracking duration in the ATC dataset, when evaluating CLiFF-LHMP, we set prediction time step $\Delta t$ to \SI{1}{\second}, CLiFF-map resolution to \SI{1}{\metre}, sampling radius $r_s$ to \SI{1}{\metre} and kernel parameter $\beta$ to 1. For training vanilla LSTM model, we set the dimension of hidden state of the LSTM model set to 128 and the learning rate set to 0.003.

For the evaluation of the predictive performance we use the \emph{Average} and \emph{Final Displacement Errors} (ADE and FDE) metrics.
%and \emph{Top-k ADE/FDE}.
ADE describes the error between points on the predicted trajectories and the respective ground truth at the same time step. FDE describes the error at the last prediction time step. %\emph{Top-k ADE} and \emph{Top-k FDE} compute the displacements between the ground truth position and the closest of $k$ predicted trajectories. For each ground truth trajectory we generate $k=5$ prediction trajectories.

We stop predicting when the sample reaches an area outside of the MoD, in case of the CLiFF map, i.e. when no SWGMMs are available within the radius $r_s$ around the sampled location. Predicted trajectories that end before $T_s$ will only be included in the ADE/FDE evaluation up to the last predicted point. When predicting for each ground truth trajectory, the prediction horizon $T_s$ is set either equal to its length or \SI{60}{\second} for longer trajectories.

\subsection{Experiments and Results}

\subsubsection{Efficiency of motion prediction with limited data}
To evaluate the data efficiency of the CLiFF-LHMP method, we ran a series of experiments with varying amount of training data [100, 200, ..., 1000 trajectories]. The training data were randomly selected multiple times. Once selected, we fed the same data to train the CLiFF-map and the LSTM model, and the evaluation metrics were averaged from all the runs.

%The training data is randomly selected from October 24, 2012, the first day of ATC dataset. The same data is used to train different models. 

\begin{figure}
\centering
\includegraphics[width=.48\linewidth]{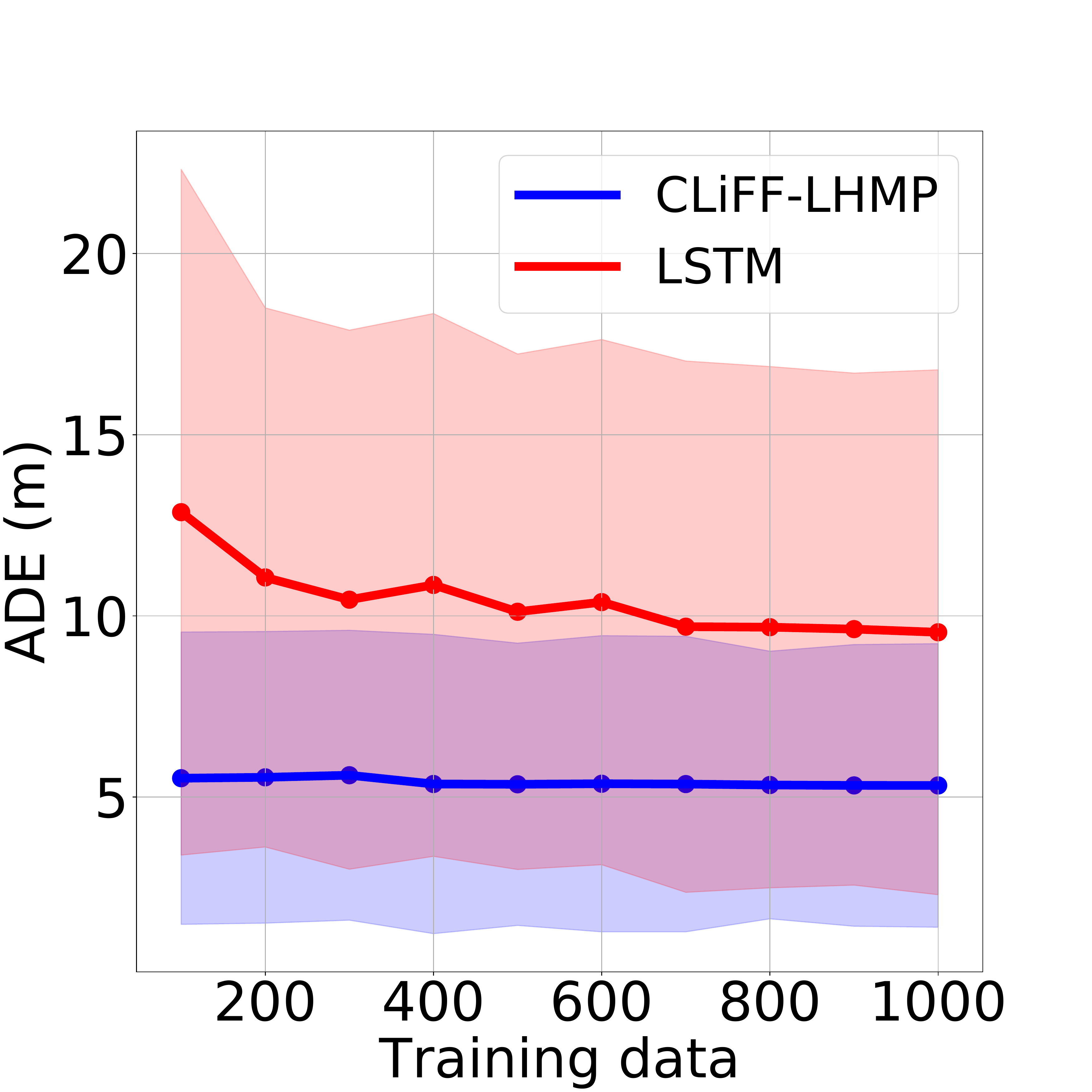}
\includegraphics[width=.48\linewidth]{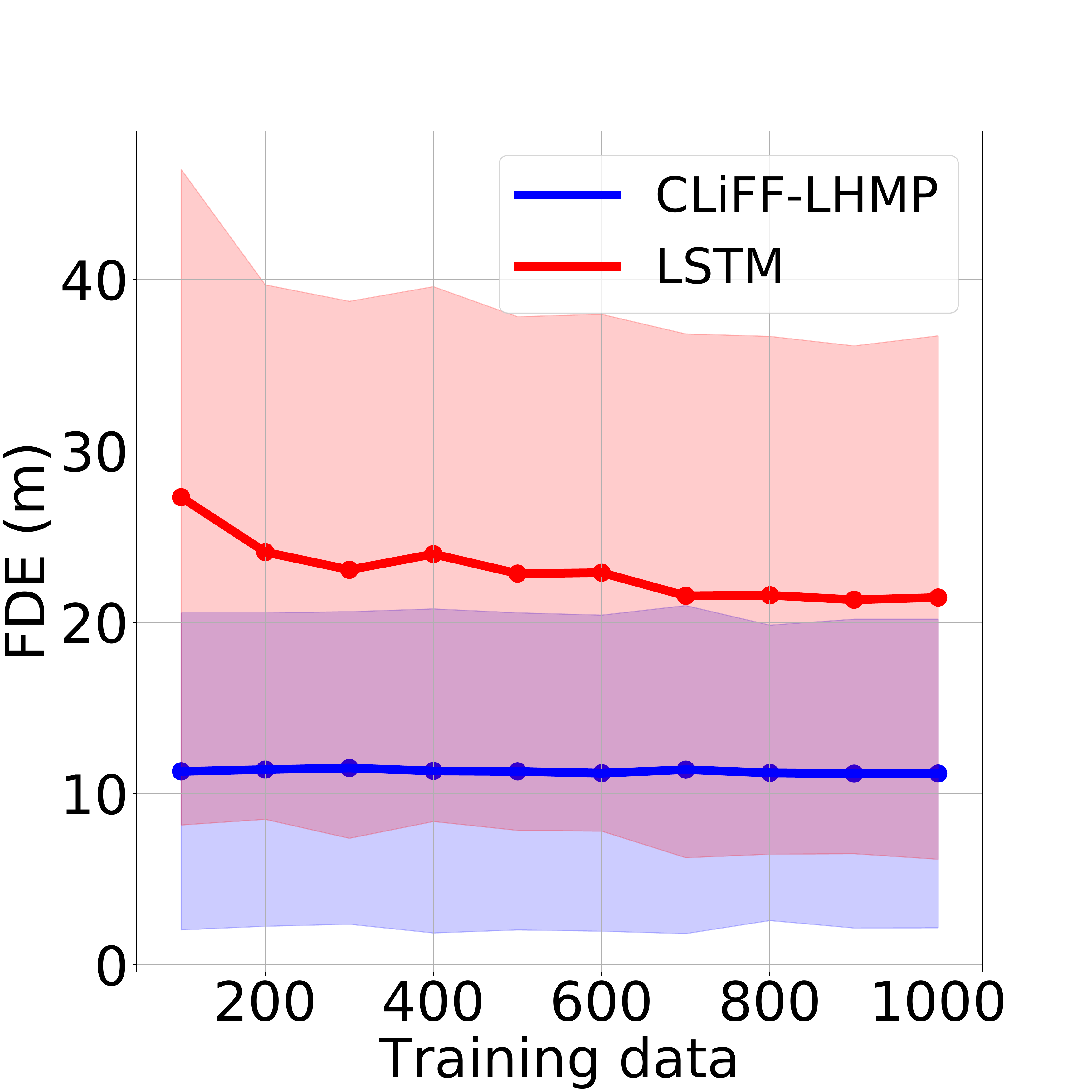}
\caption{ADE/FDE of CLiFF-LHMP and LSTM in the ATC dataset, using different amounts of trajectories (100--1000) as training data. The prediction horizon is \SI{60}{\second}. The shade represents one std. dev.}
\label{fig:both_std}
\vspace*{-3mm}
\end{figure}

\begin{figure}
\centering
\includegraphics[width=.445\linewidth]{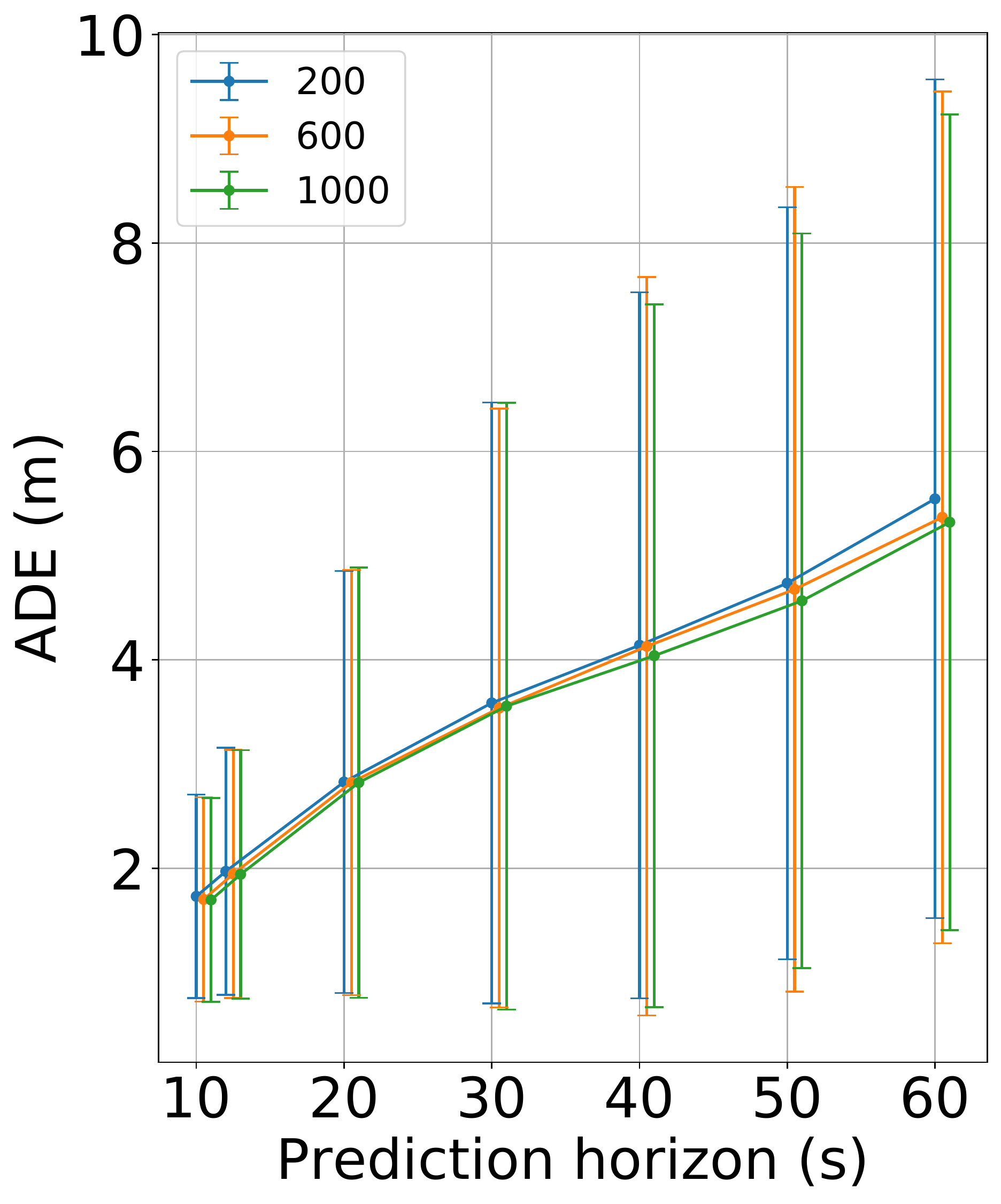}
\includegraphics[width=.445\linewidth]{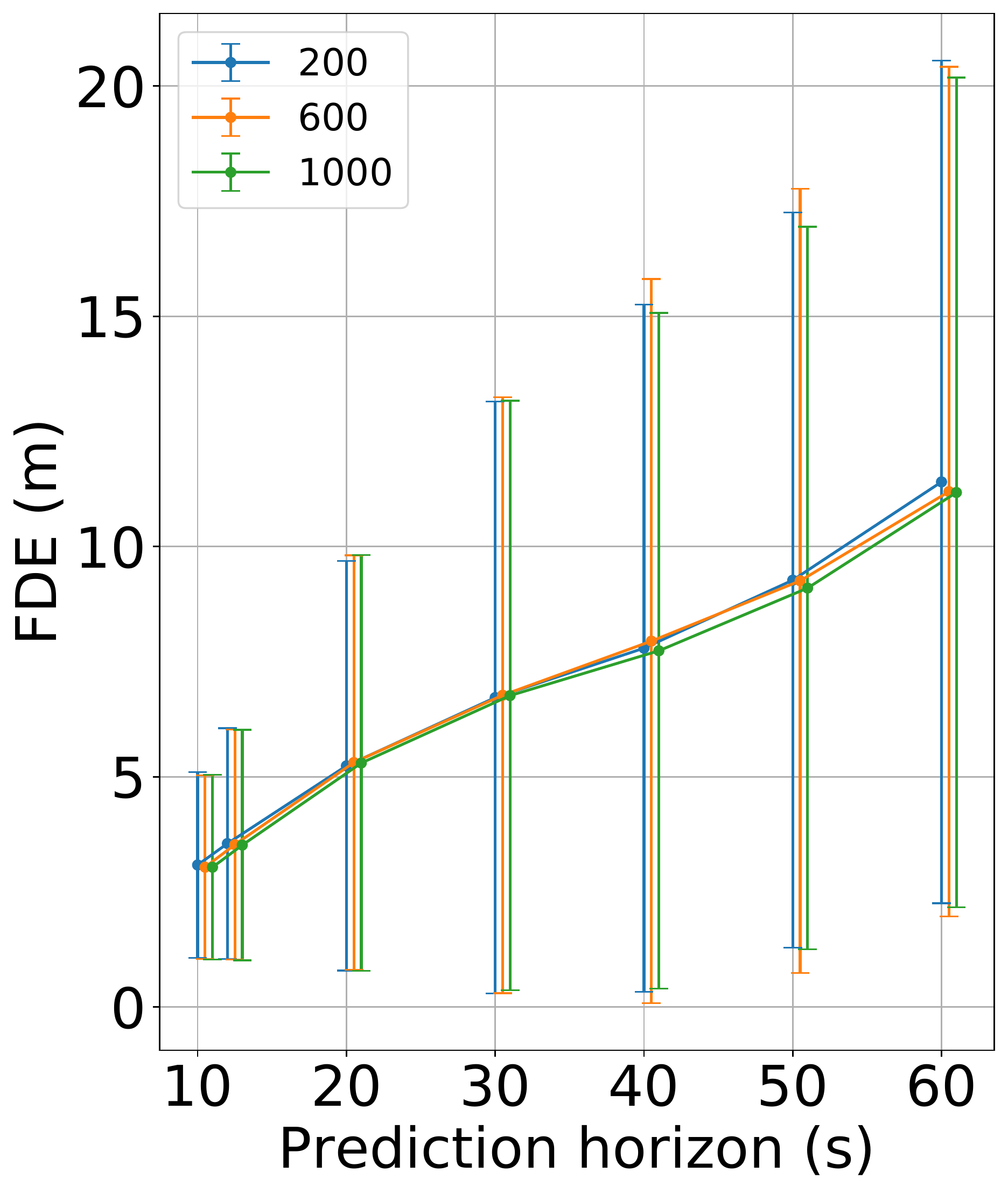}
\caption{ADE/FDE of CLiFF-LHMP in the ATC dataset with training dataset of 200, 600, 1000 trajectories and with prediction horizon 10--\SI{60}{\second}, and \SI{12}{\second}.}
\label{fig:cliff_std}
\vspace*{-5mm}
\end{figure}

\Cref{fig:both_std} shows the ADE and FDE results for CLiFF-LHMP and vanilla LSTM for prediction horizon of \SI{60}{\second}, with the number of training trajectories ranging from 100 to 1000. CLiFF-LHMP consistently outperforms LSTM when predicting long-term human motion in these cases. When more than 200 training trajectories are used, the standard deviation of ADE and FDE of CLiFF-LHMP is also lower than for LSTM. While the performance of LSTM drops substantially for smaller training data sets, especially when training with fewer than 200 trajectories, CLiFF-LHMP has a stable performance even with as few as 100 training trajectories. When decreasing the training dataset size from 1000 to 100 trajectories, the error merely increases 4\% in ADE and 1\% in FDE for CLiFF-LHMP, while for LSTM the ADE increases by 35\% and the FDE by 27\%. \Cref{fig:cliff_std} shows a comparison on different prediction horizons from \SI{10}{\second} to \SI{60}{\second} for three sizes of the training dataset (200, 600, 1000 trajectories). When the prediction horizon increases, CLiFF-LHMP becomes slightly more sensitive to the amount of training data.

%\begin{figure}
%\centering
%\includegraphics[width=.48\linewidth]{figures/workshop/CLiFF-ADE-60s.pdf}
%\includegraphics[width=.48\linewidth]{figures/workshop/CLiFF-FDE-60s.pdf}
%\caption{ADE/FDE of CLiFF-LHMP in the ATC dataset with training data 100--1000, prediction horizon \SI{60}{\second}.}
%\label{fig:cliff_line}
%\vspace*{-3mm}
%\end{figure}

%\begin{figure}
%\centering
%\includegraphics[width=.9\linewidth]{figures/workshop/1000-100-new.pdf}
%\caption{Heatmap of KL divergence of two CLiFF-maps built with 1000 and 100 training data.}
%\label{fig:kl-one}
%\vspace*{-3mm}
%\end{figure}

%\begin{figure*}[t]

%\end{figure*}

\begin{figure*}[t]
\centering
\includegraphics[clip,trim=  0mm 0mm 20mm 0mm,height=22mm]{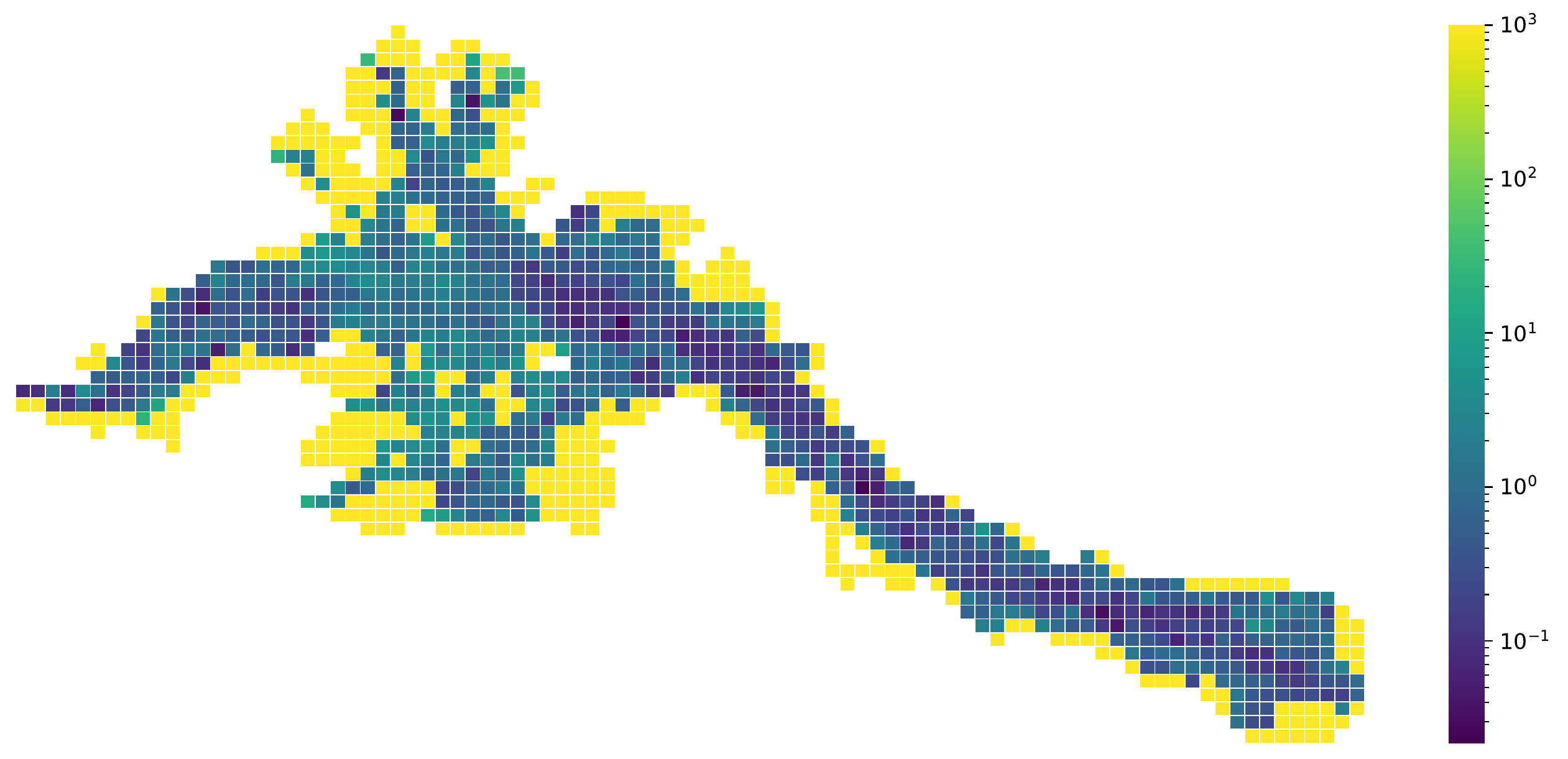}%
\includegraphics[clip,trim=  0mm 0mm 20mm 0mm,height=22mm]{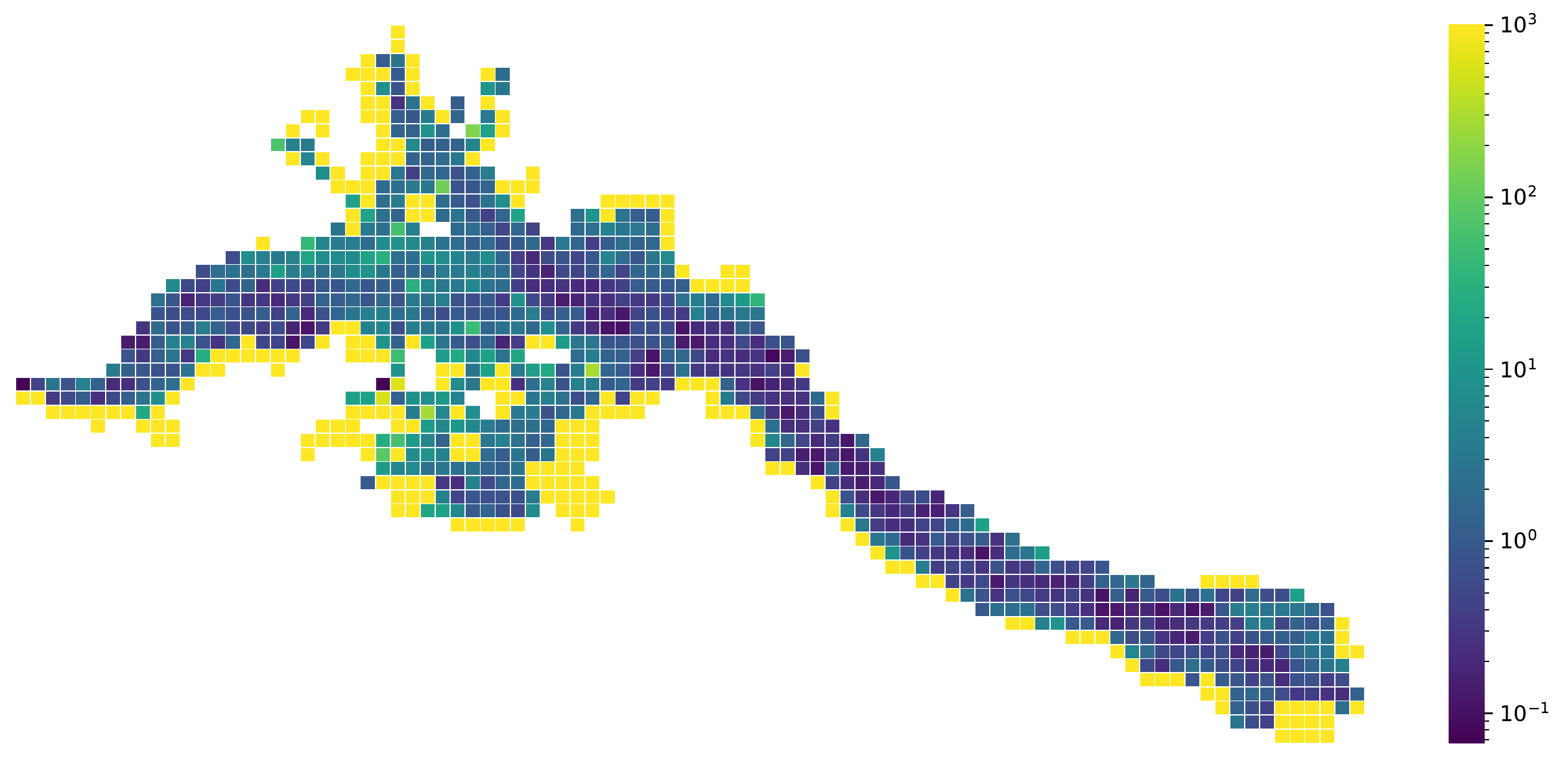}%
\includegraphics[clip,trim=  0mm 0mm 20mm 0mm,height=22mm]{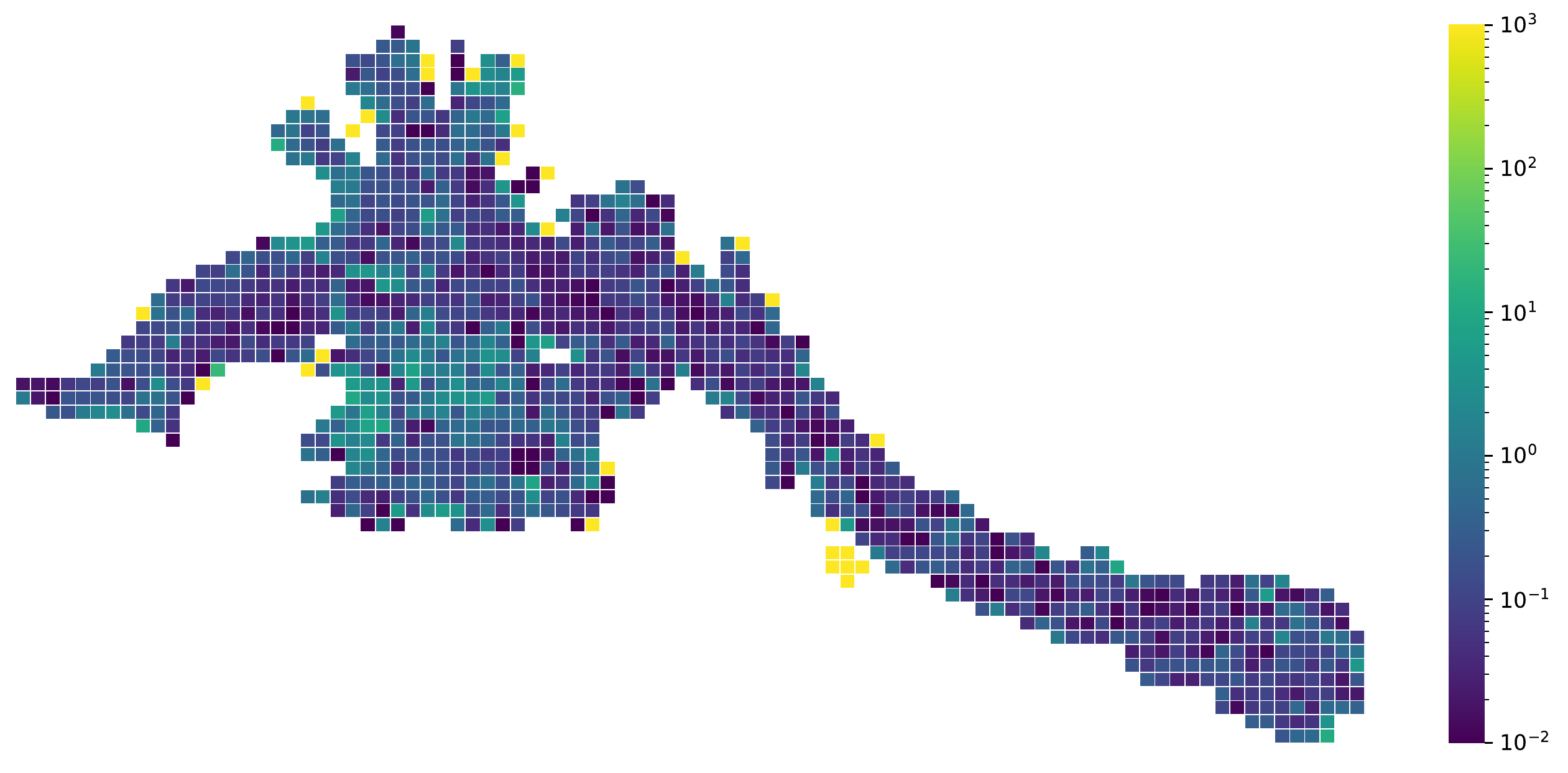}%
\includegraphics[clip,trim=  0mm  0mm 20mm 0mm,height=22mm]{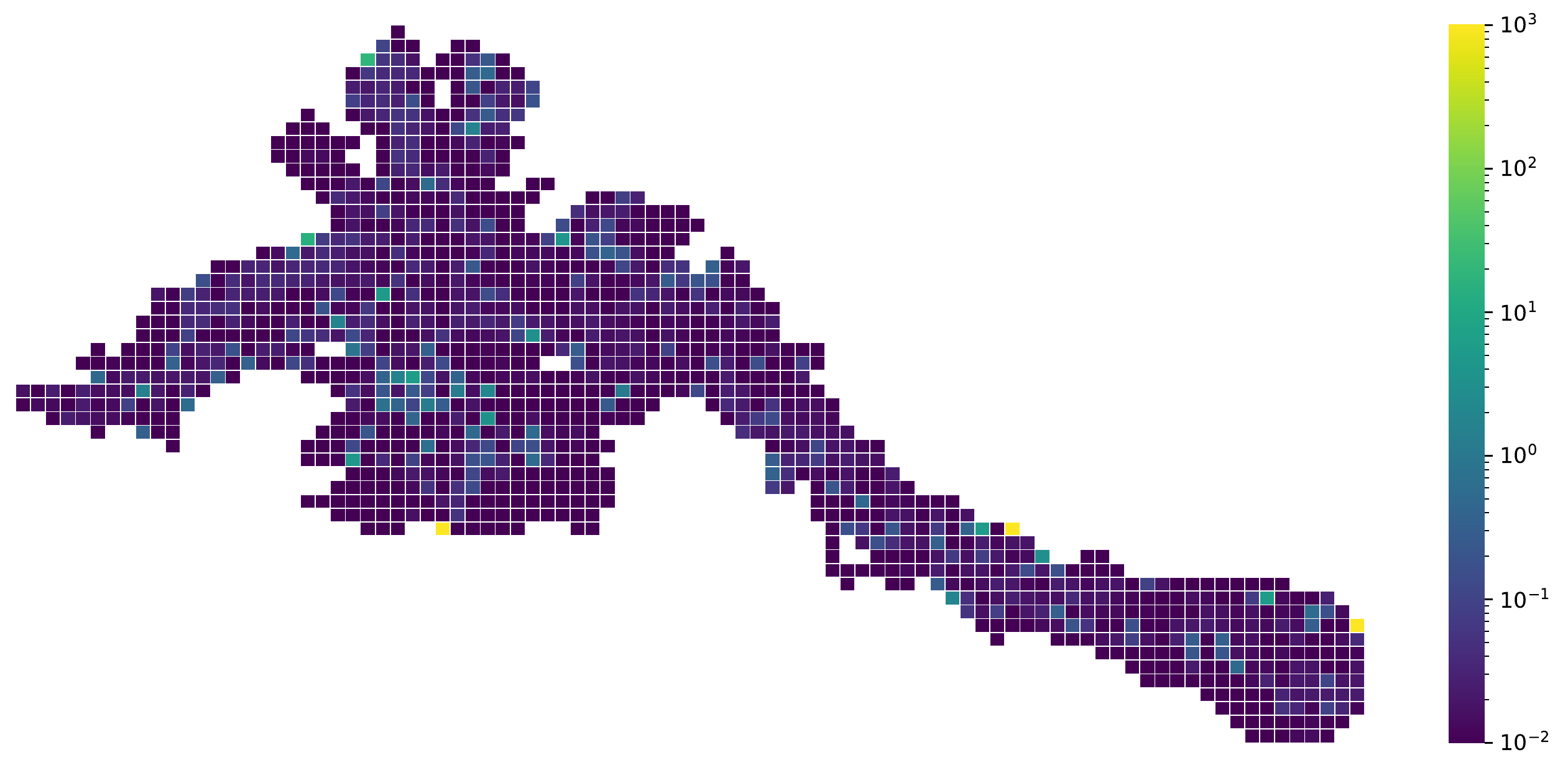}%
\includegraphics[clip,trim=  240mm  2mm 3mm 2mm,height=25mm]{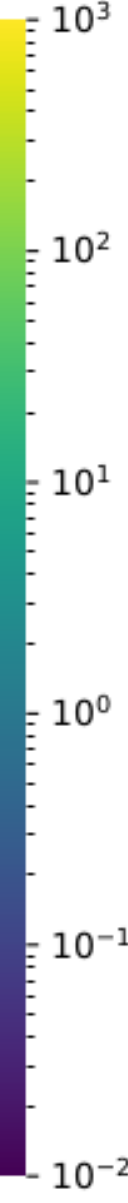}%
%\multirow{2}{*}{
%}

\caption{Heatmap of KL divergence of three pairs of CLiFF-map snapshots built with different amount of training data. \textbf{Left:} KL divergence of between CLiFF-maps trained with 1000 and 100 trajectories, \textbf{Middle left:} 200 and 100 trajectories, \textbf{Middle right:} 600 and 500 trajectories, and \textbf{Right:} 1000 and 900 trajectories. The color scale indicates the magnitude of KL divergence, with warmer colors indicating higher values, meaning larger differences between the CLiFF-map pair. When the size of training dataset increases, the impact on the CLiFF-map model becomes less significant, showing that the sensitivity of the CLiFF-map to the amount of training data decreases.}
\label{fig:kl}
\vspace*{-3mm}
\end{figure*}

\begin{figure}[t]
\centering
\includegraphics[clip,trim= 0mm 10mm 0mm 0mm,width=80mm]{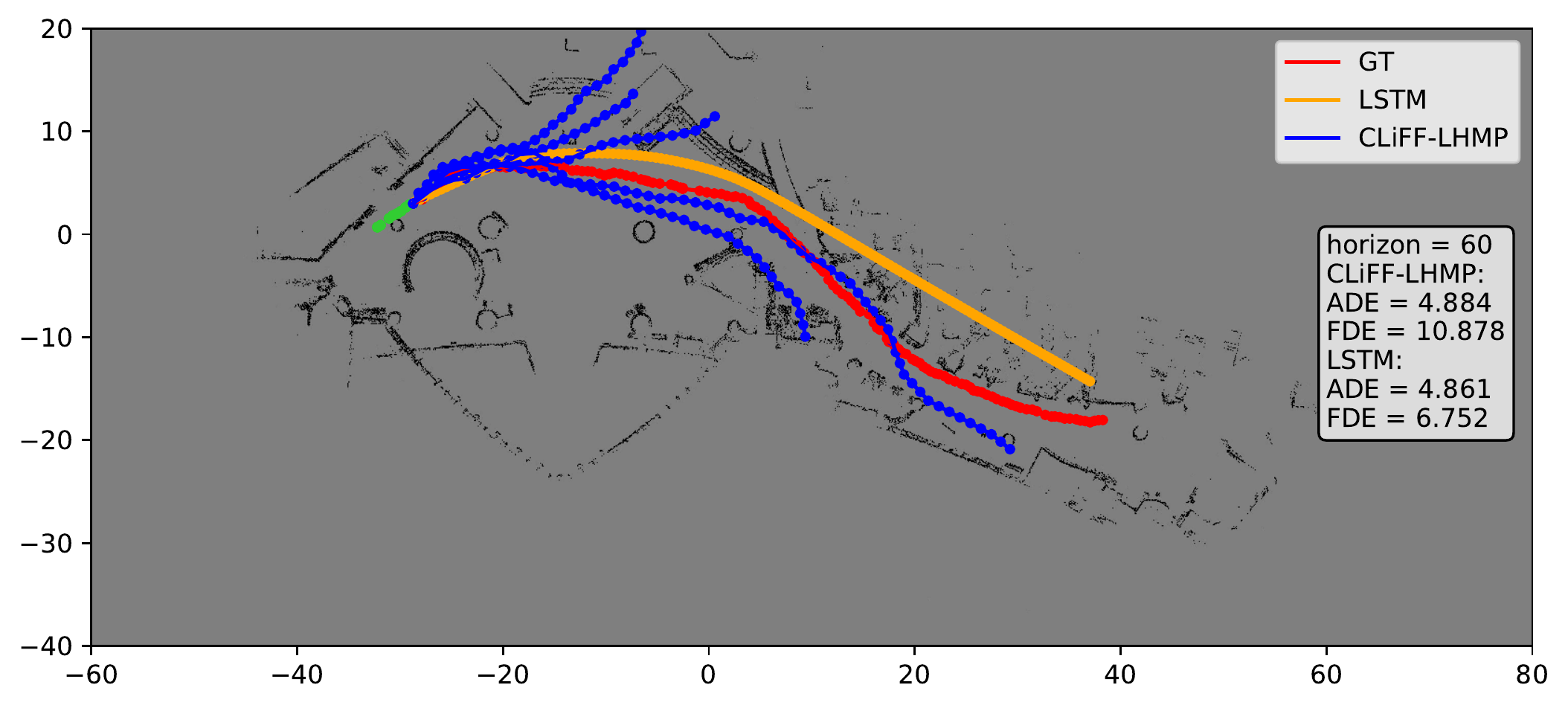}
\includegraphics[clip,trim= 0mm 10mm 0mm 0mm,width=80mm]{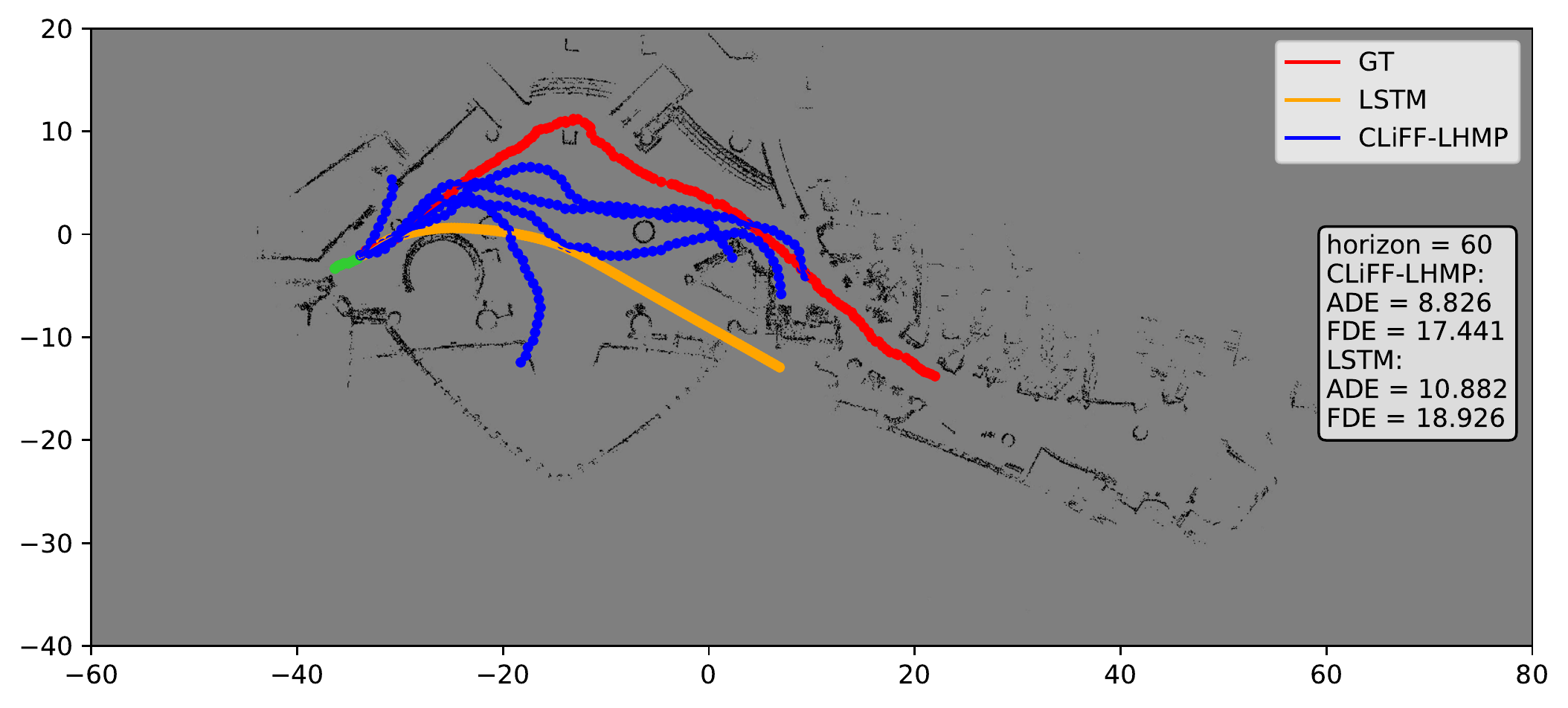}
\includegraphics[clip,trim= 0mm 0mm 0mm 0mm,width=80mm]{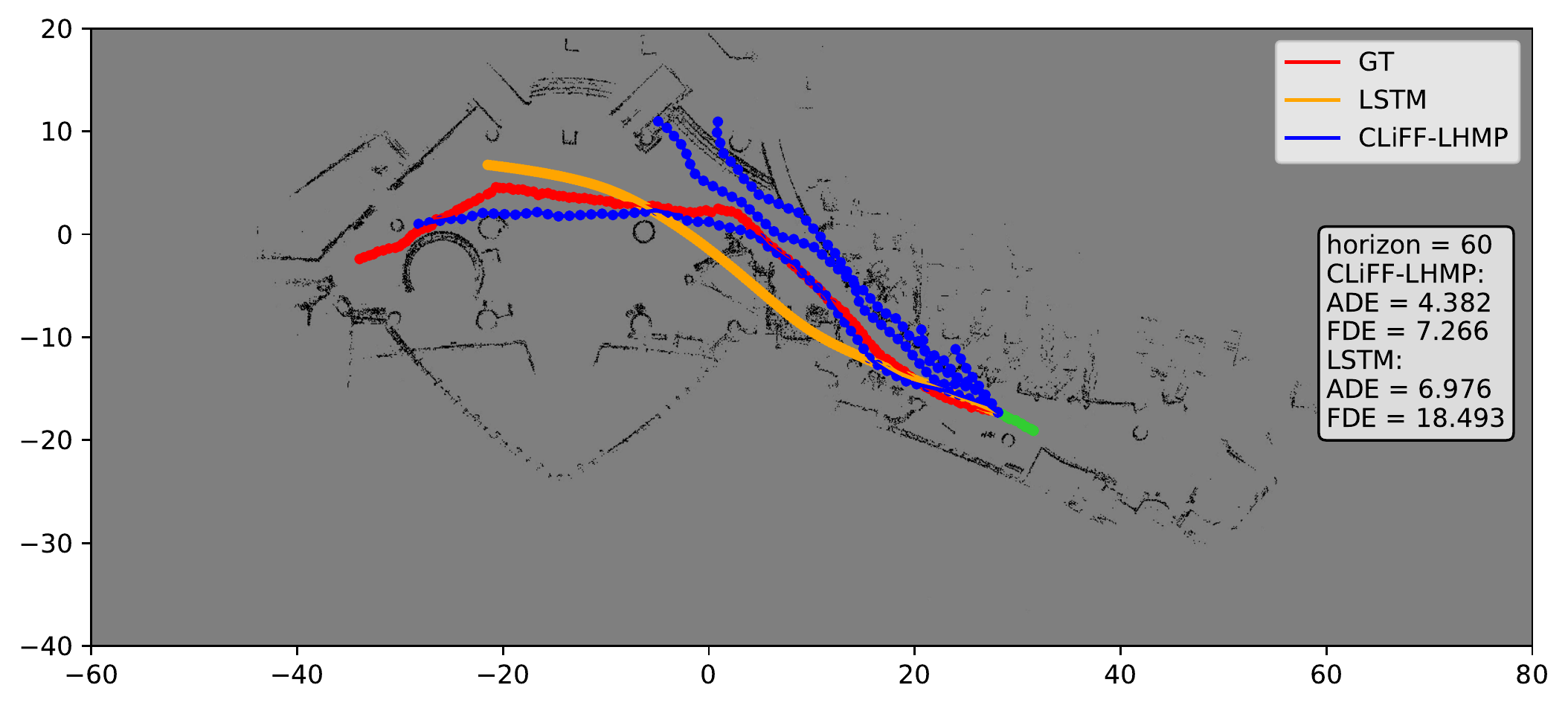}
\caption{Predictions in ATC with $T_s=60$ \SI{}{\second}. \textbf{Red} lines show the ground truth trajectory and \textbf{green} line show the observed tracklet. Prediction trajectories of CLiFF-LHMP and LSTM approaches are shown in \textbf{blue} and \textbf{orange}, respectively. When the trajectory predicted by LSTM is unfeasible by crossing the walls, CLiFF-LHMP make predictions along the corridor.}
\label{fig:atc_res}
\vspace*{-5mm}
\end{figure}

\subsubsection{Efficiency of motion representation}
To compare the quality of the underlying CLiFF-map itself, trained with different amounts of data, we compute the Kullback-Leibler (KL) divergence \cite{kullback1951information} between the distributions represented in the CLiFF-maps. The KL divergence results are shown as heatmaps in \Cref{fig:kl}. CLiFF-map associates a Gaussian Mixture Model to each location, and we use a KL divergence heatmap to visualize the changes between two different CLiFF-maps. The first image in \Cref{fig:kl} shows the changes of CLiFF-maps built with 100 and 1000 trajectories, respectively. It is evident that as the number of training trajectories increases, the primary alterations in the CLiFF-map occur predominantly along the boundary regions. Moreover, in highly constrained environments, such as the eastern corridor of the ATC map, the velocity distributions exhibit comparatively minimal variations. The other three figures in \Cref{fig:kl} shows the sensitivity of CLiFF-map to the input data. When the number of training data increases from 900 to 1000 (see the fourth image in \Cref{fig:kl}), the CLiFF-map changes less than when the number of training data increases from 100 to 200 (see the second image in \Cref{fig:kl}). This shows that the CLiFF-map can capture major human motion patterns already with small amounts of training data. 
%and update slower when we feed more data to build the MoD.

\subsubsection{Descriptive power of compact motion representation models}
%Figures~\ref{fig:atc_res}--\ref{fig:atc_3} show 
\Cref{fig:atc_res} shows qualitative examples of predicted trajectories using Maps of Dynamics in the long-term perspective. As no explicit knowledge is given about obstacle layout, LSTM predicts unfeasible trajectory which crosses the walls. In contrast, by exploiting learned motion patterns encoded in the CLiFF-map, our method predicts realistic trajectories that follow the complex topology of the environment e.g. navigating around corners or obstacles or passing through narrow passages such as doors, stairs (in the top part of the map) and exits (in the left part).

%For instance, when the trajectory predicted by LSTM is unfeasible by crossing the walls, CLiFF-LHMP make predictions along the corridor.
%Note that we correctly predict trajectories crossing obstacles such as stairs (top of the map) and exits (left of the map).

\section{CONCLUSIONS} \label{section-conclusions}
In this paper, we present the idea to exploit \emph{Maps of Dynamics} (MoDs) for long-term human motion prediction. As a proof of concept for MoD-LHMP, we propose CLiFF-LHMP. Our method uses the CLiFF-map, a specific MoD that probabilistically represents human motion patterns within a velocity field. Our approach involves sampling velocities from the CLiFF-map to bias constant velocity predictions, generating stochastic trajectory predictions for up to \SI{60}{\second} into the future. We evaluate CLiFF-LHMP using the ATC dataset, with a vanilla LSTM as the baseline approach. The experiments highlight the data efficiency advantage of our method. CLiFF-LHMP accuracy is only affected to a minor degree when using less than 200 trajectories as training data, while LSTM requires about three times as many trajectories to approach its optimal performance. The results also demonstrate that our approach consistently outperforms the LSTM method at the long prediction horizon of \SI{60}{\second}. By exploiting learned motion patterns encoded in the CLiFF-map, our method implicitly accounts for the obstacle layouts and predicts trajectories that follow the complex topology of the environment.

The current implementation of MoD-LHMP uses spatial motion patterns that are built offline based on past observations. In the future we plan to extend the approach to online life-long learning, enabling live updates based on the motion observations. One future direction is the evaluation of additional types of MoDs for long-term human motion prediction, including those capturing temporally-conditioned motion patterns. Another future direction is to learn MoDs online for life-long learning enabling updates based on live motion observations. Additionally, in future work, we aim to formally describe and analyze the MoD-LHMP methodology, include further datasets \cite{majecka2009statistical, rudenko2020thor} in the evaluation.

\printbibliography

\end{document}